\documentclass[letterpaper, 10 pt, conference]{ieeeconf}  

\IEEEoverridecommandlockouts                              

\title{\LARGE \bf
GEAR: Gaze-Enabled Human-Robot Collaborative Assembly
}

\author{Asad Ali Shahid$^{1}$, Angelo Moroncelli $^{1}$, Drazen Brscic $^{2}$, Takayuki Kanda $^{2}$, and Loris Roveda$^{1}$ 
\thanks{$^{1}$Asad Ali Shahid, Angelo Moroncelli and Loris Roveda are with SUPSI, IDSIA, Lugano, Switzerland. 
        {\tt\small asadali.shahid@supsi.ch}}%
\thanks{$^{2}$Drazen Brscic and Takayuki Kanda are with the Graduate School of Informatics, Kyoto University, Japan.}%
\thanks{This work was partially supported by the project n. IC00I0L-231370 \textit{A toolkit for designing human-centered collaborative robotic assembly system}, funded from SNSF. This work was partially supported by the European Union, Grant Agreement n. 101093126 (project ACES: Autopoiesis Cognitive Edge-cloud Services) and by the Swiss State Secretariat for Education, Research and Innovation (SERI) under contract number 22.00490.}
}

\usepackage{cite}
\usepackage{amsmath,amssymb,amsfonts}
\usepackage{algorithmic}
\usepackage{graphicx}
\usepackage{textcomp}
\usepackage{xcolor}
\usepackage{booktabs}
\usepackage{caption}
\usepackage{comment}
\usepackage{hyperref}
\def\BibTeX{{\rm B\kern-.05em{\sc i\kern-.025em b}\kern-.08em
    T\kern-.1667em\lower.7ex\hbox{E}\kern-.125emX}}
\begin{document}


\maketitle


\maketitle

\begin{abstract}
Recent progress in robot autonomy and safety has significantly improved human-robot interactions, enabling robots to work alongside humans on various tasks. However, complex assembly tasks still present significant challenges due to inherent task variability and the need for precise operations. This work explores deploying robots in an assistive role for such tasks, where the robot assists by fetching parts while the skilled worker provides high-level guidance and performs the assembly. We introduce GEAR, a gaze-enabled system designed to enhance human-robot collaboration by allowing robots to respond to the user's gaze. We evaluate GEAR against a touch-based interface where users interact with the robot through a touchscreen. The experimental study involved 30 participants working on two distinct assembly scenarios of varying complexity. Results demonstrated that GEAR enabled participants to accomplish the assembly with reduced physical demand and effort compared to the touchscreen interface, especially for complex tasks, maintaining great performance, and receiving objects effectively. Participants also reported enhanced user experience while performing assembly tasks. Project page:\href{https://sites.google.com/view/gear-hri}{sites.google.com/view/gear-hri}
\end{abstract}


\section{Introduction}
In modern manufacturing, the assembly tasks involve numerous parts requiring precise operations across a wide range of product variants. Neither humans nor robots can efficiently complete the entire assembly process on their own. A collaborative approach leverages the complementary strengths of each partner: human dexterity and intelligence combined with robot precision. Such human-robot collaboration can enhance productivity, improve task quality, and reduce ergonomic strain on workers. 

For robots to effectively assist in assembly tasks, they need to accurately infer the user's intentions. For example, if a robot anticipates that the user will assemble a specific part in the next step, it can proactively retrieve and deliver the part, minimizing the user's idle time. In human communication, social cues, such as eye gaze, are fundamental for signaling engagement, interest, and attention. During manipulation tasks, individuals naturally focus their gaze on the object of interest before making a corresponding movement, typically shifting their gaze to the target object approximately 600 milliseconds before grasping it \cite{land2001ways}. Similarly, in collaborative settings, such as object assembly, team members use gaze to indicate which tool or component they need from their teammates \cite{wang2019gaze}. Eye gaze can therefore serve as a natural and intuitive signal for conveying a user’s intent to robots, potentially enabling more seamless collaboration.

\begin{figure}[t!]
    \vspace{2mm}
    \centering \includegraphics[width=0.94\columnwidth]{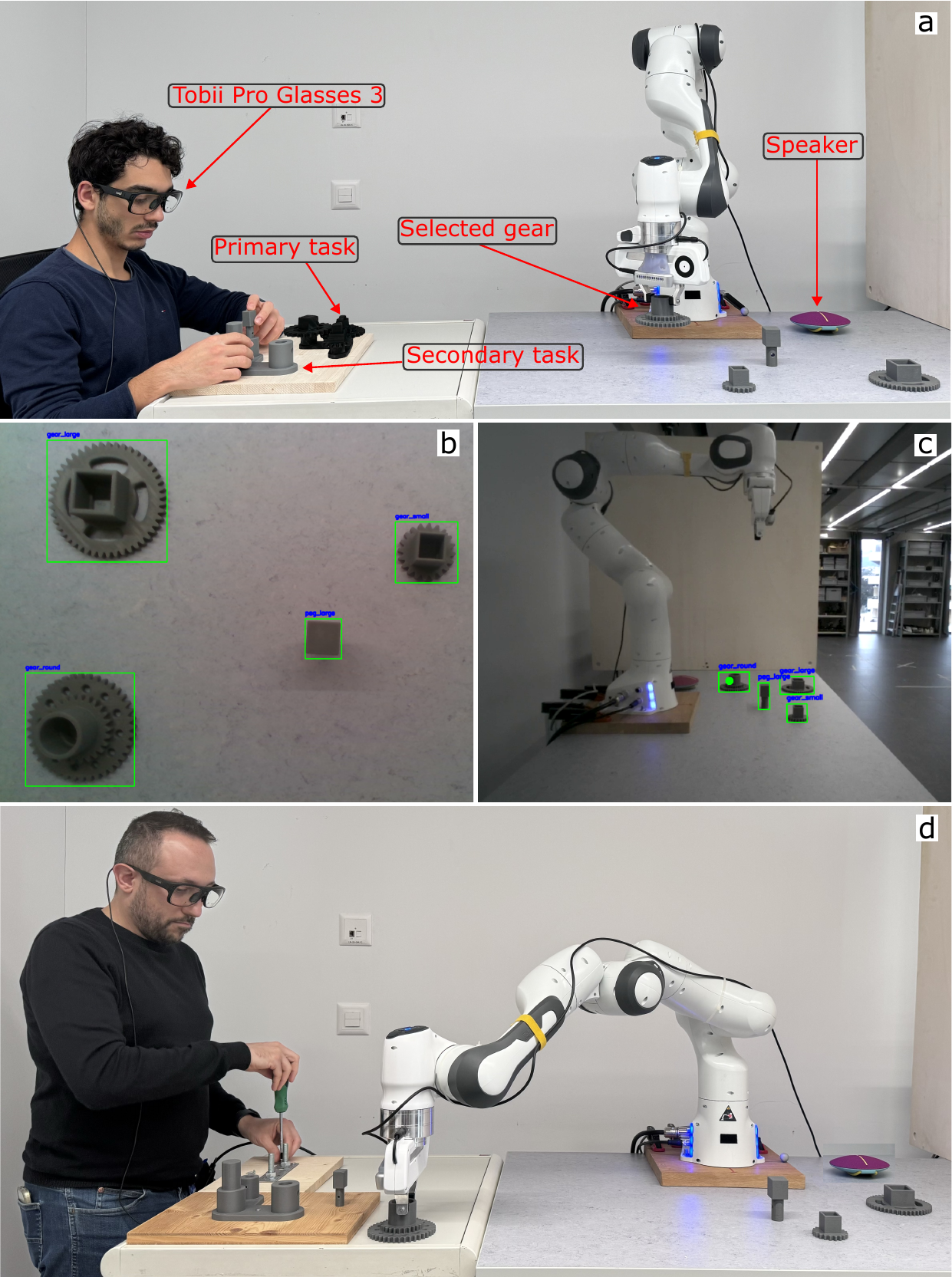}
    \caption{GEAR enables users to request desired parts from the robot simply by looking at them while working on a primary task. The user's intention is inferred by integrating their eye gaze with object detection on both the robot's side (Fig. b) and the user's side (Fig. c). The system is evaluated in two distinct collaborative assembly scenarios: assembling two sets of gears (Fig. a), and fastening screws while assembling a gear set (Fig. d).}
    \vspace{-4mm}
    \label{fig:overview}
\end{figure}

\vspace{-1mm}

While gaze is promising for human-robot interaction (HRI), the investigation of gaze-based systems in collaborative assembly tasks is limited. This is partly due to the current limitations of gaze tracking technology and the challenges in effectively translating gaze information into robot actions. In particular, the 3D gaze estimation accuracy of most eye trackers is too low to achieve reliable robot control. For instance, if the robot is tasked with delivering parts to a user during an assembly based on their intention, it must be able to accurately determine the position of the parts. Although converting 3D gaze coordinates into the robot's reference frame could theoretically indicate object locations, most eye trackers lack the precision for effective manipulation.

To address these limitations, we introduce GEAR (see Fig. \ref{fig:overview}), a human-in-the-loop system that integrates gaze tracking and object detection to enhance human-robot collaboration in assembly tasks. The human provides a higher-level guidance by directing their gaze at a specific object, while the robot autonomously brings the object once it detects that the human is gazing at it. The system also incorporates auditory feedback to confirm the robot’s actions and guide users through the correct assembly sequence, ensuring smoother interaction. GEAR allows users to maintain focus on the primary task while intuitively requesting assistance from the robot. To evaluate the system's efficacy, we conducted a user study with 30 participants, comparing the usability and performance of the gaze-based interface with that of a touch-screen-based interface.

In summary, the main contributions of this work are as follows:

\begin{itemize}
    \item We develop a gaze-guided interaction system for human-robot collaborative assembly, combining gaze tracking with object detection alignment to enable intuitive, real-time collaboration without explicit commands.
    \item We conduct a comprehensive user study comparing the usability and performance of our gaze-based system with a touch-based system, where users request objects via a touchscreen interface, evaluating both systems in terms of user workload, user satisfaction, and task efficiency in a collaborative assembly setting.

\end{itemize}

\section{Related Works}

\subsection{Collaborative Assembly Systems}

Recent advances in human-robot collaboration have shifted from robots executing isolated sub-tasks \cite{matheson2019human, michalos2018seamless} to intelligent partners that leverage perception and learning algorithms \cite{unhelkar2018human, nemlekar2023transfer, cai2024hierarchical}. Several frameworks have been proposed for collaborative assembly—addressing task allocation \cite{tsarouchi2017human, johannsmeier2016hierarchical}, scheduling \cite{Wilcox2012OptimizationOT}, and quality control \cite{banh2015inspector}. In contrast to these approaches, our work specifically focuses on integrating gaze-based interactions into collaborative assembly systems.

\subsection{Communication Interfaces}
Prior research in HRI has explored various communication modalities to convey human intentions to robots. Gesture-based interfaces, for instance, have shown significant potential, enhancing communication and improving task efficiency \cite{sheikholeslami2017cooperative, mazhar2019real}. Gestures have been used to streamline tasks like vehicle door assembly \cite{gleeson2013gestures}. On the other hand, auditory signals, such as speech-based communication \cite{chai2014collaborative}, can also be effective but often struggle in noisy environments. 

Eyes tend to move rapidly and serve both as a means of gathering and conveying information. This makes eye-tracking an effective way to convey the intentions to other agents \cite{huang2016anticipatory, palinko2016robot}. Prior studies have shown that gaze cues can improve shared attention and fluency during handovers \cite{moon2014meet}, and provide contextual cues that enhance the robot's understanding of human intentions \cite{huang2016anticipatory}. Eye gaze tracking has also been used to enhance imitation learning in robotic manipulation by discarding task-irrelevant objects \cite{kim2020using} and to assess the operator's comfort level during a collaboration \cite{dufour2020visual}. In the realm of assistive robotics, gaze has been used to improve robotic rehabilitation, reducing workload for the user \cite{shafti2019gaze, padmanabha2024independence}. The authors in \cite{aronson2018eye} proposed a system that enhances user input via a joystick with autonomous functionalities of the robot, effectively inferring user intentions through eye gaze tracking. 

Despite advancements in gaze-based collaborative systems, leveraging eye gaze for collaborative assembly remains notably limited. While gaze has been used to infer user intention in prior research, there is a lack of studies utilizing gaze for real-time parts retrieval in human-robot assembly systems. This study addresses this gap by developing and evaluating a gaze-guided interaction system that integrates eye gaze tracking with object detection on both the robot and the user side, enabling intuitive communication during assembly tasks.

\begin{figure*}[t!]
    \vspace{2mm}
    \centering \includegraphics[width=0.88\textwidth]{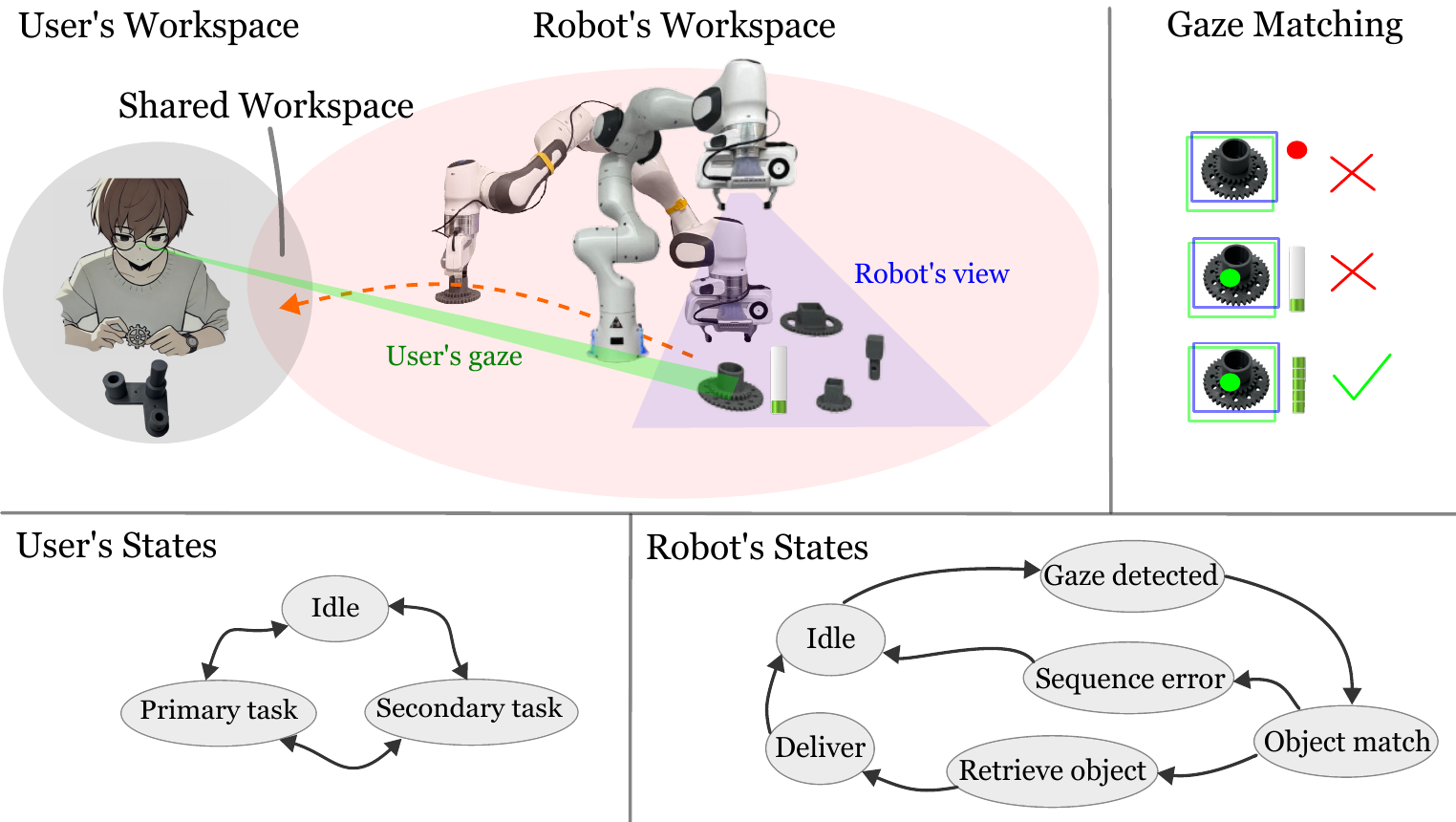}
    \caption{Overview of the GEAR system in collaborative assembly, where the user focuses on a primary task while fixating on a desired part. The robot utilizes its object detection along with the user's gaze to accurately localize the part before retrieving it and delivering it to the user. Gaze matching is triggered only when the user maintains a steady gaze on a part, indicating intent for physical interaction, rather than with brief or transient glances. A simplified state machine at the bottom shows the key states for both the user and the robot.}
    \label{fig:approach}
\end{figure*}

\section{System Design and Implementation}
The purpose of this work is to develop a robotic system that leverages human gaze patterns and environmental context to enhance collaborative assembly tasks. We envision a scenario where a skilled user collaborates with a robot, which assists by fetching parts needed for the next assembly step based on the user's gaze. Previous studies used the robot's own sensors to estimate human gaze for the object stacking task \cite{palinko2016robot}, but this requires the human to face the robot, significantly limiting its practical use. Modern eye-tracking devices offer 3D gaze tracking via head-mounted wearables, yet their accuracy is often insufficient for reliable robot control. In this work, we employ 2D gaze matching and object detection on both the user’s and robot’s sides to accurately interpret the user's intentions and translate them into effective robot commands. Figure \ref{fig:approach} depicts an overview of our GEAR system.  

\subsection{Eye Tracking}
For eye-tracking, we use Tobii Pro Glasses 3\footnote{https://www.tobii.com/products/eye-trackers/wearables/tobii-pro-glasses-3}, a wearable eye-tracker designed to be worn like regular glasses (see Figure \ref{fig:overview}). These glasses are equipped with multiple sensors and cameras to measure and record eye movements. They use infrared light to illuminate the eyes and capture reflections from the cornea and pupil in an eye image, which is then processed to determine the user's point of gaze \cite{tobii}. A front-facing scene camera records the user’s field of view, capturing what the user is looking at in real-time. The glasses require a one-time calibration for each user to compute a 3D eye model for optimizing gaze estimation. During calibration, the user is instructed to focus on a bullseye marker for a few seconds. Successful calibration is achieved when the user's gaze overlaps with the bullseye on the calibration card. 


\subsection{Gaze Matching and Object Alignment}
Gaze matching in the context of HRC involves correlating gaze data with specific targets to understand about user's attention. One approach to this is determining whether a user's 2D gaze coordinates fall within the bounding boxes of identified objects in a visual scene. We adopt this method because it provides a straightforward and computationally efficient way to track and analyze user attention in real-time and assess whether the user intends to interact with a particular object.

The user’s gaze is captured using an eye-tracker, which provides 2D gaze coordinates $\mathit{(x_g, y_g)}$. We compute the mean gaze $\mathit{(\bar{x}_g, \bar{y}_g)}$ of the past 15 sample points, corresponding to approximately 0.6 to 0.8 seconds, to reduce noise and improve accuracy. This helps filter out transient glances, ensuring that only sustained attention is considered indicative of the user’s potential intention to interact with an object. A user may glance at multiple objects to identify which one they are interested in, without necessarily intending to interact. Simultaneously, an object detection algorithm processes the input images, identifying objects and producing bounding boxes defined by their corner coordinates $\mathit{(x_{min}, y_{min})}$ and $\mathit{(x_{max}, y_{max})}$. The gaze matching process can then be represented as:

\begin{equation}
    \text{Gaze\_Match} = \begin{cases}
        1 & \text{if } x_{\text{min}} \leq \bar{x}_g \leq x_{\text{max}} \\
          & \quad \text{and } y_{\text{min}} \leq \bar{y}_g \leq y_{\text{max}} \\ 
        0 & \text{otherwise}
    \end{cases}
\end{equation}

\normalsize

For object detection, we used Mask R-CNN \cite{he2017mask} pretrained on the Microsoft Coco dataset. Since there are no industrial assets in this dataset, we fine-tuned the pre-trained model on 40 real-world images of the parts that were used in our assembly task (pegs and gears). 

To perform object alignment, the robot verifies that it detects the user’s intended object by checking the labels of components within its field of view. If the user's target object is identified, the robot estimates its position to retrieve it. Note that the same object detector is used for both the user's and the robot's views. Before reaching for the desired part, the robot plays a sound stating, “Object X selected; Bringing now,” to alert the user about its action. Furthermore, the assembly sequence is encoded within the robot's system; if the user’s desired object is visible but requires additional parts to be assembled first, the robot will provide audio feedback to inform the user of the necessary steps to take before proceeding. To play sound, the Google Text-to-Speech (gTTS) library\footnote{https://gtts.readthedocs.io/en/latest/} is employed to convert text into audio. This allows the robot to vocalize messages clearly, providing audible feedback to users regarding its actions and status during the task. 

\subsection{Description of Touch Screen Interface}
\begin{figure}[htbp]
    \centering \includegraphics[width=0.8\columnwidth]{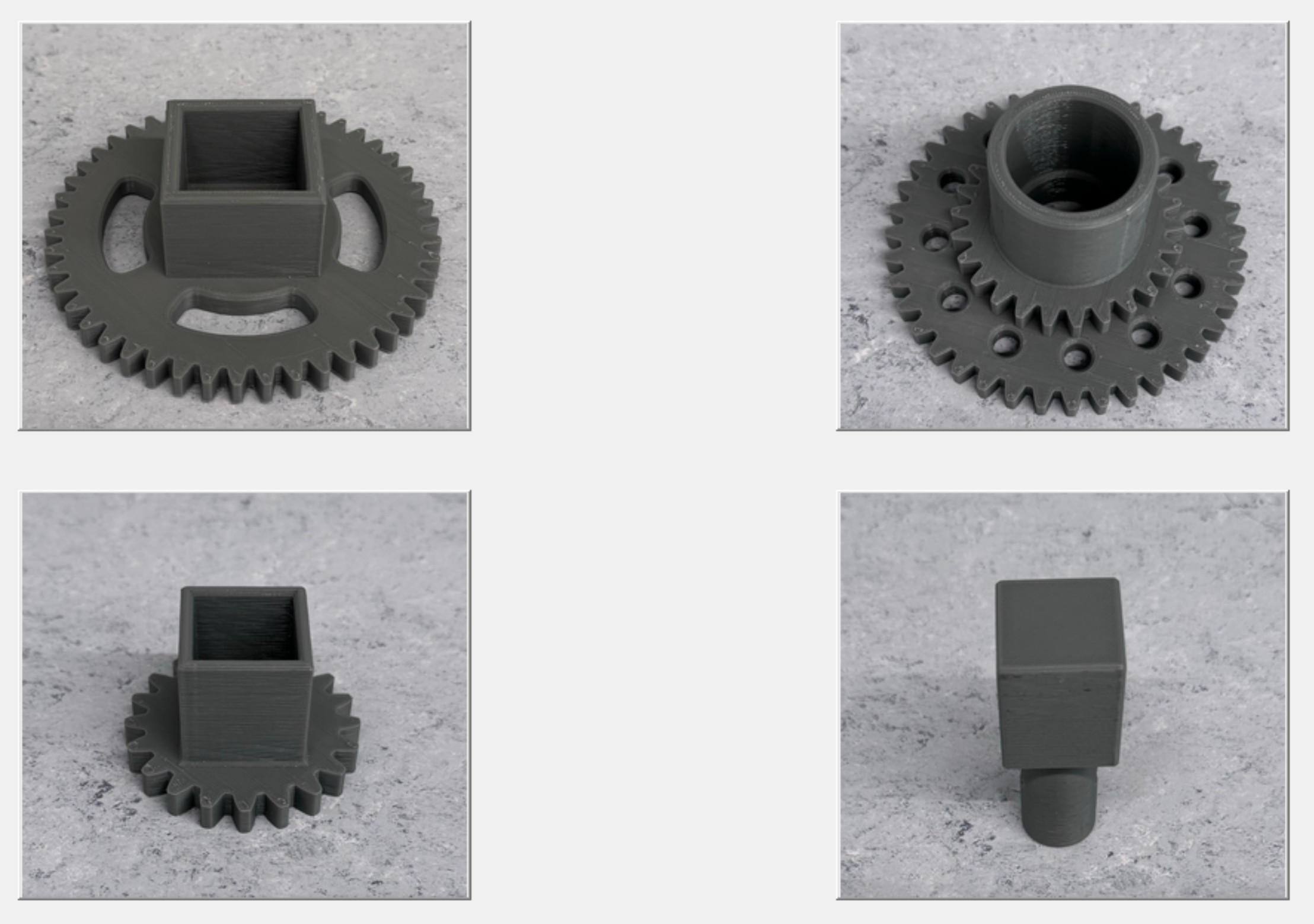}
    \caption{Touch-screen interface used in the experiments. Users selected the desired part by tapping the corresponding image on the screen.}
    \vspace{-3mm}
    \label{fig:touch}
\end{figure}

For the baseline system, we designed a touch-screen-based interface (Figure \ref{fig:touch}), displaying images of the available parts on a screen. To request a specific part, users simply tapped on the image of the desired part, allowing the robot to retrieve it. Such touchscreen interfaces are a common standard in many industries due to their simplicity and ease of use.

\section{Experiments}

\subsection{Tasks and Environment} 
\label{sec:tasks}
We set up human-robot collaborative assembly experiments in which the robot assists by retrieving parts that the user anticipates are needed for the next assembly step. The experiments were conducted in two distinct assembly scenarios: one involving the assembly of two gear sets, and the other focusing on assembling a gear set and fastening screws on a nut-bolt assembly. In both scenarios, users had a primary task to complete, for which all necessary parts were available at their workstation. For the secondary task, however, the required parts were located farther away, and users needed to request them from the robot. The initial arrangement of these parts was randomized within the robot's workspace at the beginning of each trial to reflect variability typical in industrial conditions. Users had to manage both primary and secondary tasks in parallel. In a practical context, the primary and secondary tasks could represent different stages in a complex assembly process, such as engine assembly.

\begin{figure}[t!]
    \centering \includegraphics[width=0.98\columnwidth]{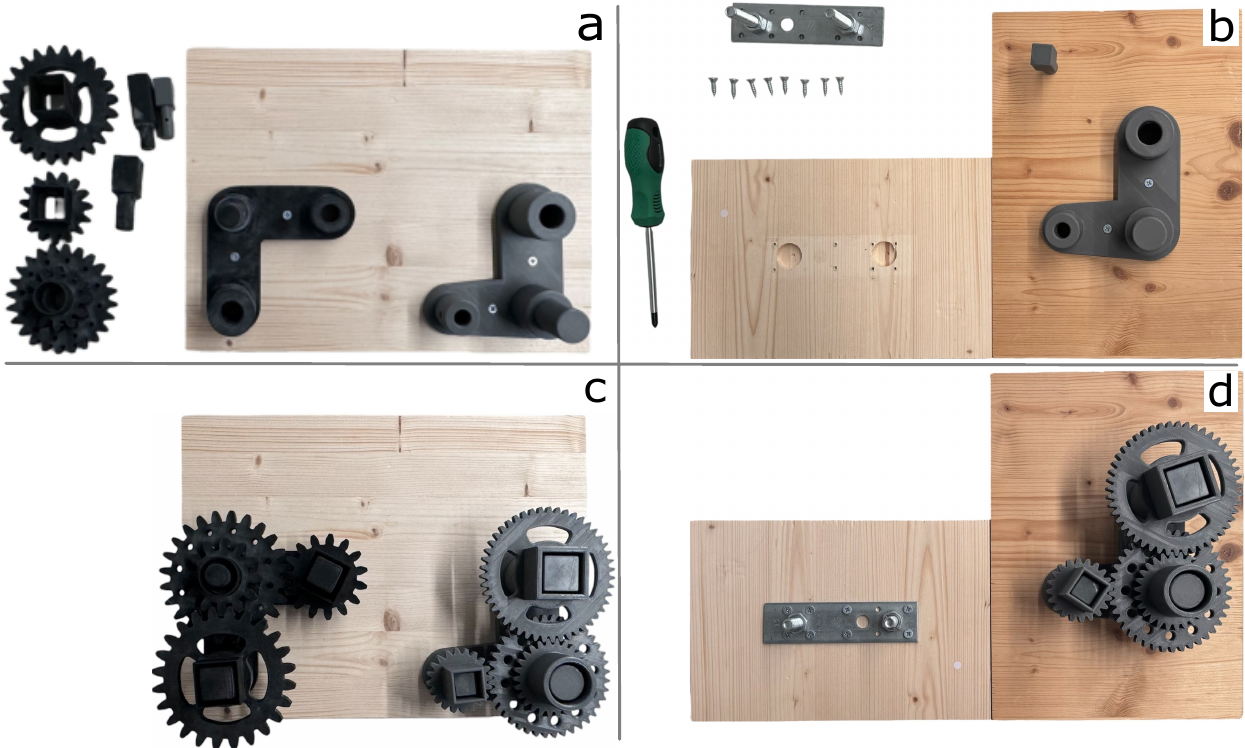}
    \caption{Initial and final configuration of two assembly scenarios: \textit{Gear Assembly} (Figs. a and c), \textit{Gear and Nut-Bolt Assembly} (Figs. b and d). Each user begins the task with a workspace setup as in the top Figures, and after completing the assembly process, the final configurations are depicted in the bottom figures.}
    \vspace{-3mm}
    \label{fig:task_config}
\end{figure}

\subsubsection{Gear Assembly} 
In the first scenario, participants assembled two gear sets, each requiring 5 components (Figure \ref{fig:task_config}c). The primary task involved assembling a black gear set with all parts placed at the workstation, while the secondary task required assembling a grey gear set, for which one component was near the user and four were placed further away and had to be retrieved by the robot. Figures \ref{fig:task_config}a and \ref{fig:task_config}c illustrate the initial setup and final configuration, respectively.

\subsubsection{Gear and Nut-Bolt Assembly} 
The second scenario combined gear assembly with screw fastening on a nut-bolt plate (Figure \ref{fig:task_config}d). The primary task focused on fastening screws using tools provided at the workstation, while the secondary task again involved assembling a grey gear set (one part near the user, four retrieved by the robot). Users managed both tasks simultaneously. Figures \ref{fig:task_config}b and \ref{fig:task_config}d show the initial setup and the final arrangement after task completion.

\begin{figure*}[htbp]
    \centering \includegraphics[width=1\textwidth]{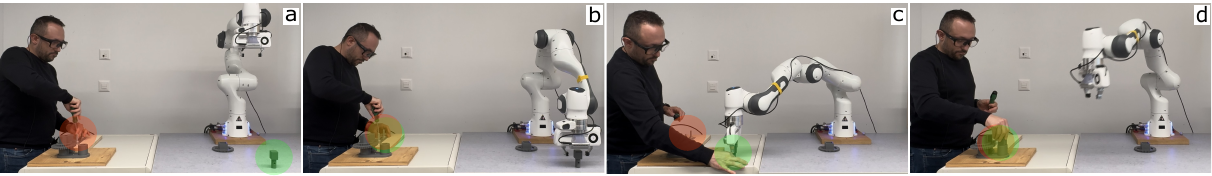}
    \caption{The sequence of frames illustrating different phases of a collaborative assembly task: the user is gazing at the desired part (green area in Fig. a) while working on a primary task, the robot approaches to retrieve it (Fig. b), robot bringing the part to the shared workspace where the user takes it (Fig. c), the user assembling the part (Fig. d).}
    \label{fig:exp_sequence}
\end{figure*}

\subsection{Participants}
This study included a total of 30 participants, comprising 21 males and 9 females, with an average age of $30 \pm 5$ years. Half of the participants had corrected vision (wearing eyeglasses, which were removed during experiments), while the others had normal vision. The participants were divided into two groups of 15, each assigned to complete one of the two distinct tasks described in Section \ref{sec:tasks}.

Group 1 consisted of 11 males and 4 females, including 6 participants with corrected vision and 9 with normal vision. Group 2 included 10 males and 5 females, with 9 having corrected vision and 6 having normal vision. Group 1 was tasked with assembling two sets of gears, whereas Group 2 engaged in a task that involved assembling a single gear set and fastening screws to secure a nut-bolt plate. All participants were proficient in English.


\subsection{Experimental Protocol}
Each participant took part in a within-subjects experiment, experiencing both the gaze-guided and touch-based systems in random order to avoid bias. Before the experiments, participants were provided informed consent after a briefing on the study’s purpose and procedure. For the gaze experiments, the eye-tracker was calibrated for each individual before the trial. Both experiments lasted 15-20 minutes, depending on the participant's task completion time and the time required to set up the system between trials. The study was approved by the ethics committee at the author's institution  (Approval No. CE\_2024\_23).

\subsection{Hypotheses}
 We hypothesize that using gaze for complex assembly tasks in human-robot collaboration can reduce the operator's workload, possibly enhancing task efficiency. The analysis aimed to assess the workload, the user experience, and the effectiveness during two distinct assembly scenarios: \textit{Gear Assembly} and \textit{Gear and Nut-Bolt Assembly}. We formulated the following hypotheses to guide our investigation:
 
\begin{enumerate}
  \item \textbf{H1}: GEAR reduces the perceived workload compared to the touch-based interface.
  \item \textbf{H2}: GEAR improves user experience compared to the touch-based interface.
  \item \textbf{H3}: GEAR helps users to complete assembly tasks more effectively than the touch-based interface when the task workload increases.
\end{enumerate}

\subsection{Evaluation Metrics}\label{AA} To assess the usability, user experience, and workload of our GEAR system, we used both subjective and objective measures. Objectively, we recorded task completion times. Subjectively, we employed two widely recognized evaluation tools: the System Usability Scale (SUS) \cite{brooke2013sus} and NASA Task Load Index (NASA-TLX) \cite{hart2006nasa}. Together, these tools provide a comprehensive assessment of performance, user experience, and workload.

\subsubsection{System Usability Scale (SUS)}
The SUS consists of 10 standardized questions that assess different aspects of the system's usability, each rated on a 5-point Likert scale, where 1 corresponds to ``strongly disagree" and 5 to ``strongly agree". The resulting scores provide an overall measure of usability. The key advantage of using SUS is that it is a widely recognized and validated tool, offering benchmarks from various studies across different technologies and interfaces. 

\subsubsection{NASA Task Load Index (NASA-TLX)}
The NASA-TLX, on the other hand, measures cognitive workload across six dimensions: Mental Demand, Physical Demand, Temporal Demand, Performance, Effort, and Frustration. By providing a multidimensional evaluation, NASA-TLX identifies the workload experienced during interactions with robotic systems, allowing for the optimization of interfaces and collaboration strategies to improve efficiency and reduce cognitive strain in human-robot collaboration settings.

Both the NASA-TLX and SUS questionnaires were administered on a smartphone app after each trial, presented in a random order to each participant. 

\subsubsection{Interview Questionnaire}
\label{sec:questionnaire}
To understand users' perception and preference regarding the interface during the task, we asked each participant three questions after they completed both trials:

\begin{itemize}
    \item Which system do you prefer, gaze-based or touch-based, and why?
    \item How difficult was it to select the desired object using gaze, on a scale of 0 (extremely easy) to 5 (extremely difficult)?
    \item How would you rate the time it took for the robot to start moving after the command was given: too long, long, appropriate, short, or too short?
\end{itemize}

\section{Results}
\subsection{Gaze Accuracy Evaluation}
\begin{figure}[t]
    \centering \includegraphics[width=0.9\columnwidth]{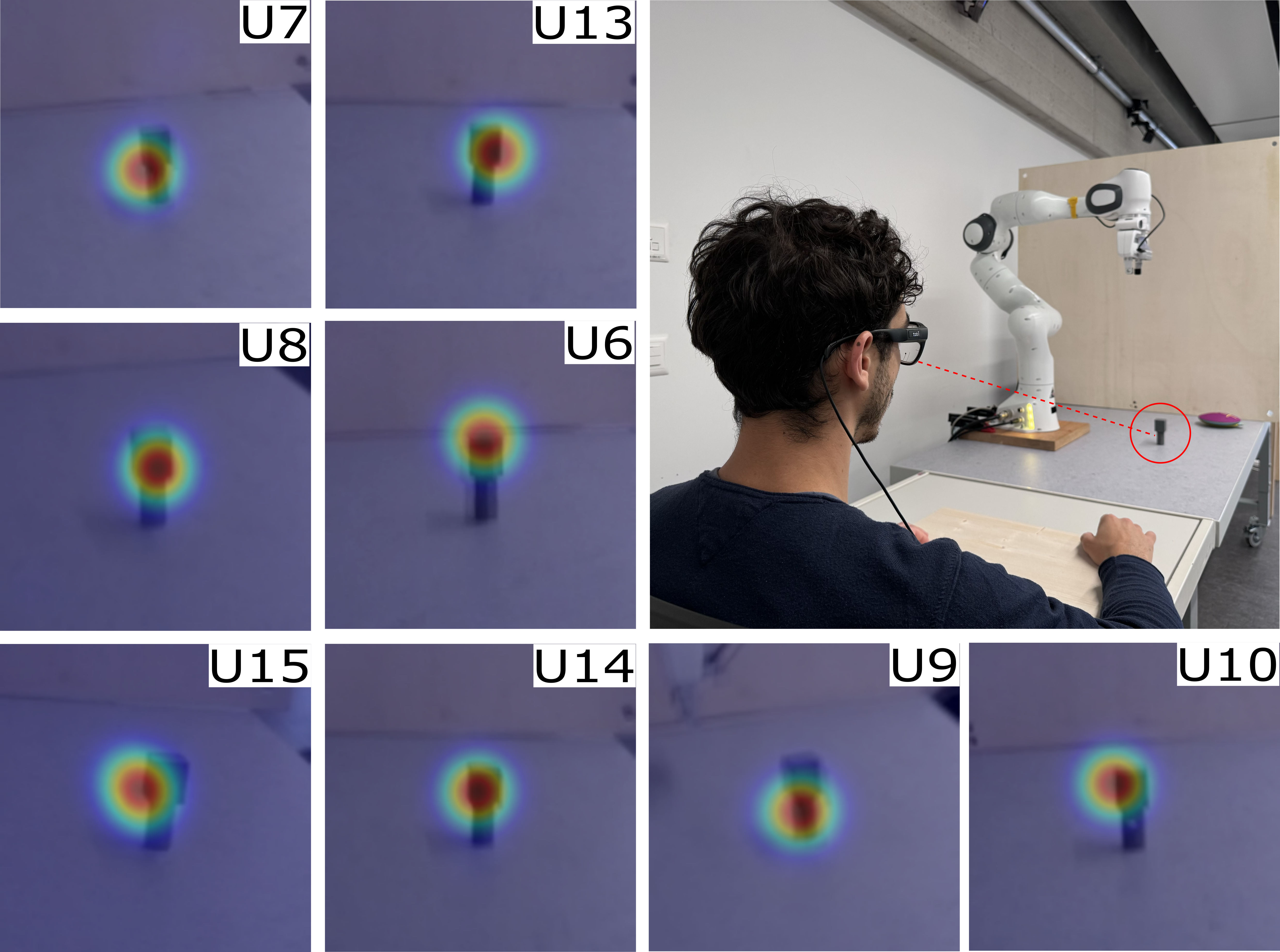}
    \caption{Accuracy of 2D gaze estimation using Tobii Pro Glasses 3. The images show results from an experiment with 15 participants, assessing gaze accuracy relative to the object center. Heatmaps represent gaze data from eight random users, the same as in Figure~\ref{fig:gaze_acc}, while the top-right image shows one participant during the experiment.}
    \label{fig:gaze_acc_hm}
\end{figure}

\begin{figure}[htbp]
    \centering \includegraphics[width=0.98\columnwidth]{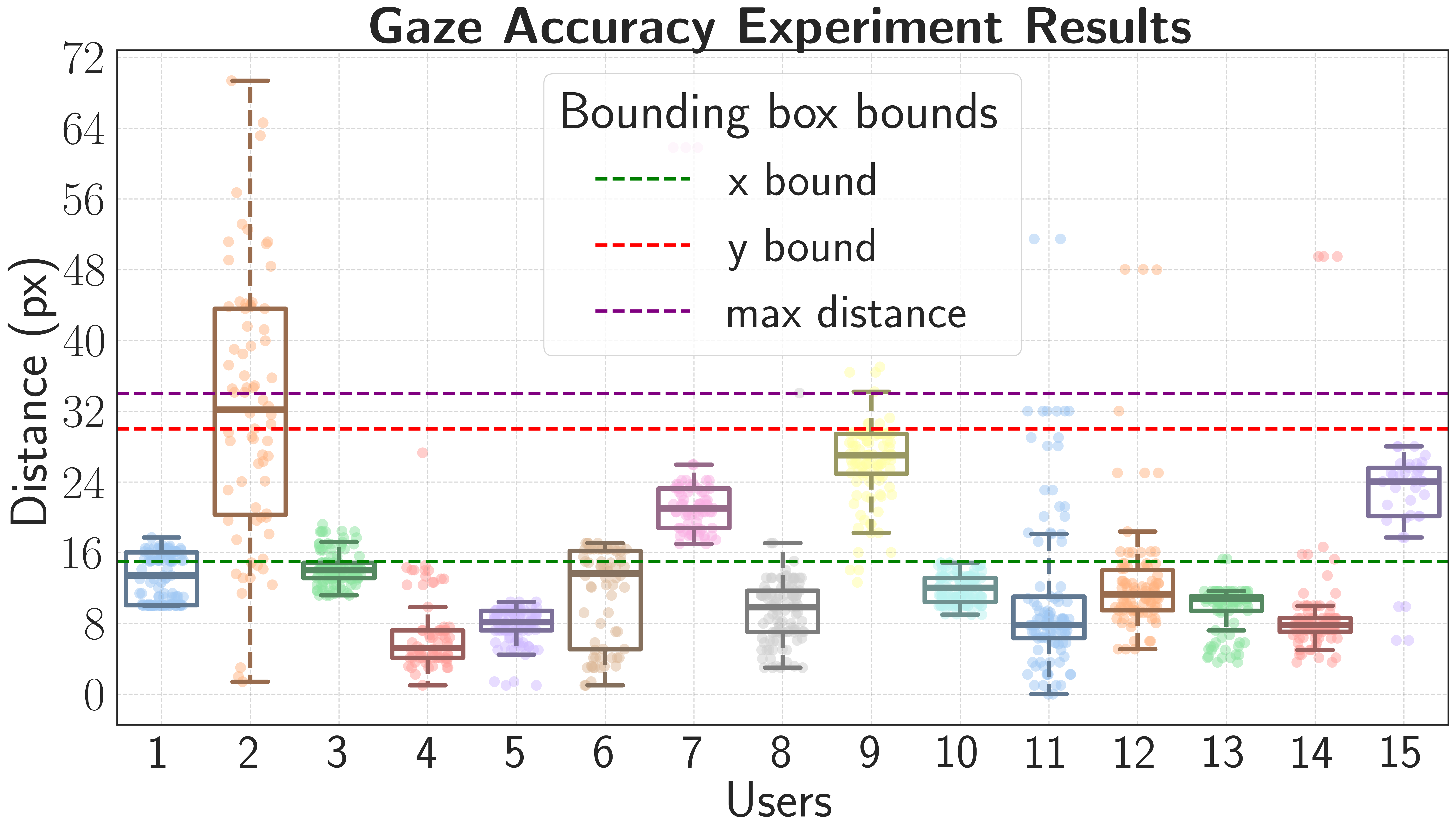}
    \caption{The plot presents results from the gaze accuracy experiment, where the Euclidean distance (x-axis) of each gaze point from the object center (used to create the heatmaps in Figure~\ref{fig:gaze_acc_hm}) is calculated. On average, the data shows that when users (y-axis) focus on the object, their gaze remains within the object's bounding box. For each user, both the raw data (individual points) and the aggregated data distribution are displayed as boxplots.}
    \vspace{-3mm}
    \label{fig:gaze_acc}
\end{figure}

To evaluate the accuracy of the eye tracker's 2D gaze estimation, we experimented with 15 randomly selected users (12 males, 3 females) from Groups 1 and 2 before starting the assembly tasks. Six participants had corrected vision, while the remaining nine had normal vision. Each participant fixated on the center of a peg (Figure~\ref{fig:gaze_acc_hm}) for 4 seconds, during which we recorded image frames and corresponding gaze samples. To remove transient effects, we discarded the first image and the initial 10 gaze samples from each trial; the remaining points were used to generate a gaze heat map over the averaged images (Figure~\ref{fig:gaze_acc_hm}). Figure~\ref{fig:gaze_acc} presents the quantitative results by plotting each user on the x-axis and the distribution of the Euclidean distance of their 2D gaze from the object center on the y-axis. The three horizontal dashed lines indicate the thresholds for gaze-based object selection, where the x and y bounds are set by the distances from the object center along the bounding box dimensions, and the maximum distance is defined as the distance from the object center to the farthest corner. This confirms that for all users, the gaze falls with high probability within the bounding box of the object

\begin{table}[b!]
\centering
\caption{Mean TLX Scores for GEAR and touchscreen interface across two assembly scenarios.}
\begin{tabular}{lcc}
\toprule
\textbf{Task} & \textbf{Touch} & \textbf{GEAR} \\
\midrule
\textit{Gear Assembly} & 24.39 & 17.39 \\
\textit{Gear and Nut-Bolt Assembly} & 26.17 & 16.28 \\
\bottomrule
\end{tabular}
\label{tab:mean_tlx}
\end{table}

\begin{figure}[h]
    \centering \includegraphics[width=0.98\columnwidth]{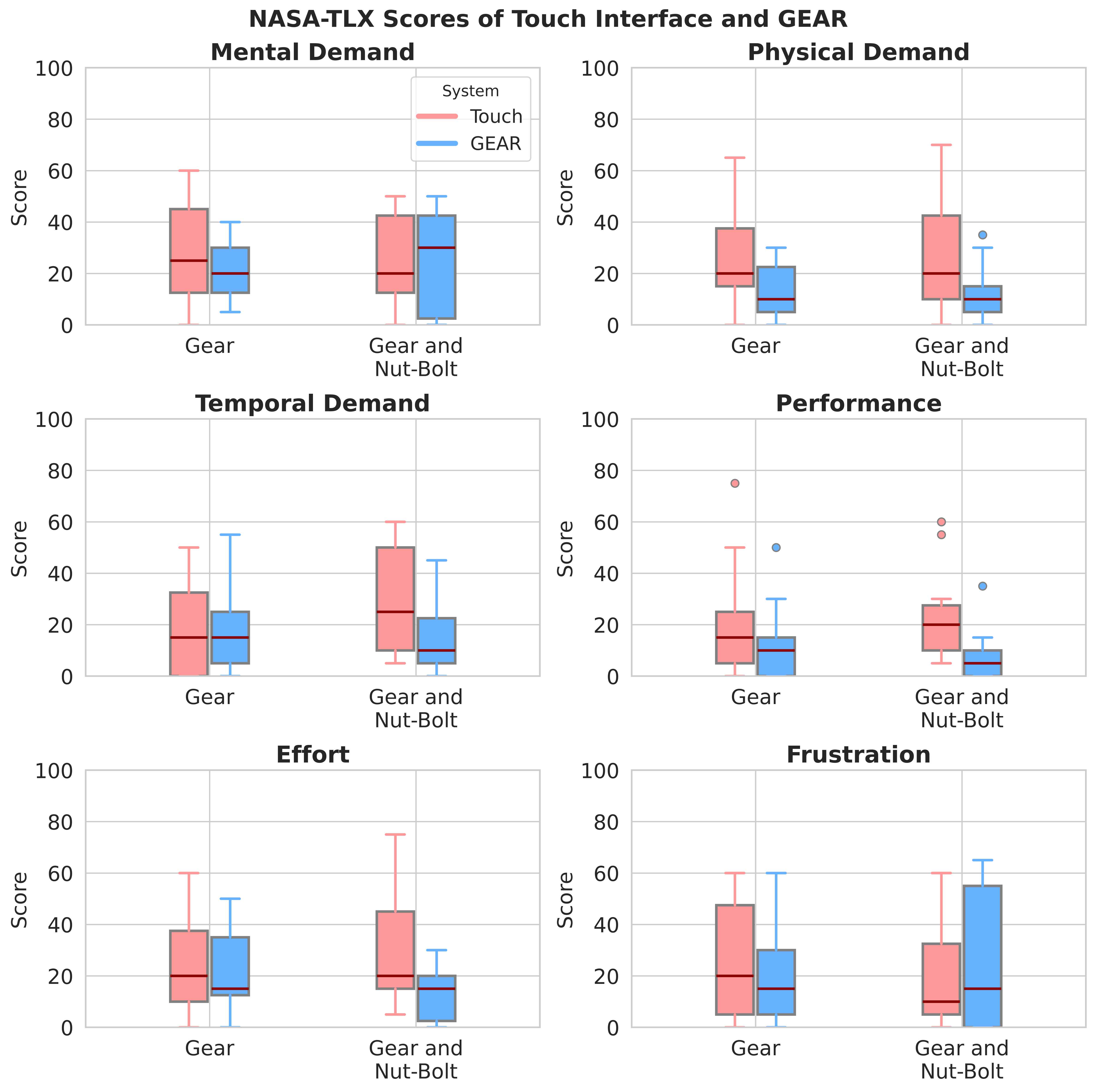}
    \caption{The plot shows the six metrics of the NASA-TLX Index, each in a separate inset. In each inset, results are compared between two groups of 15 users: one group using both the touch interface and the GEAR system during the \textit{Gear Assembly} experiment, and another group of 15 users during the \textit{Gear and Nut-Bolt Assembly} experiment. This visualization complements our statistical analysis, providing a detailed view of the results. The boxplots display the interquartile range (IQR), median line, and whiskers extending to data points within 1.5 times the IQR.}
    \label{fig:tlx}
\end{figure}

\begin{figure}[h]
    \centering \includegraphics[width=0.98\columnwidth]{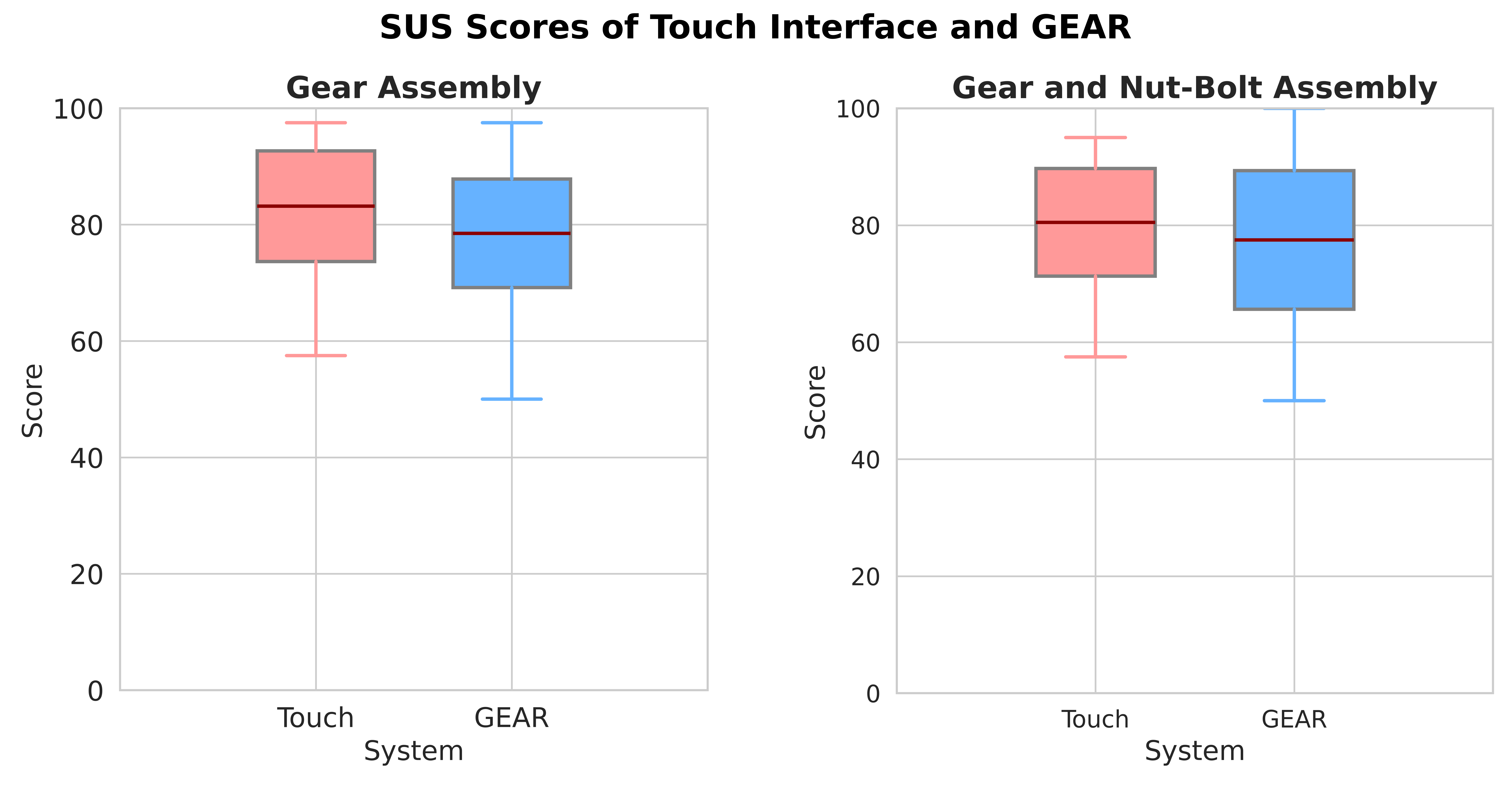}
    \caption{The figure presents two boxplots comparing the System Usability Scale (SUS) results for two experiments: \textit{Gear Assembly} and \textit{Gear and Nut-Bolt Assembly}. In each plot, we analyze the distribution of scores from 15 users who engaged with both the touch interface and GEAR.}
    \label{fig:sus}
\end{figure}

\begin{figure}[h]
    \centering \includegraphics[width=0.98\columnwidth]{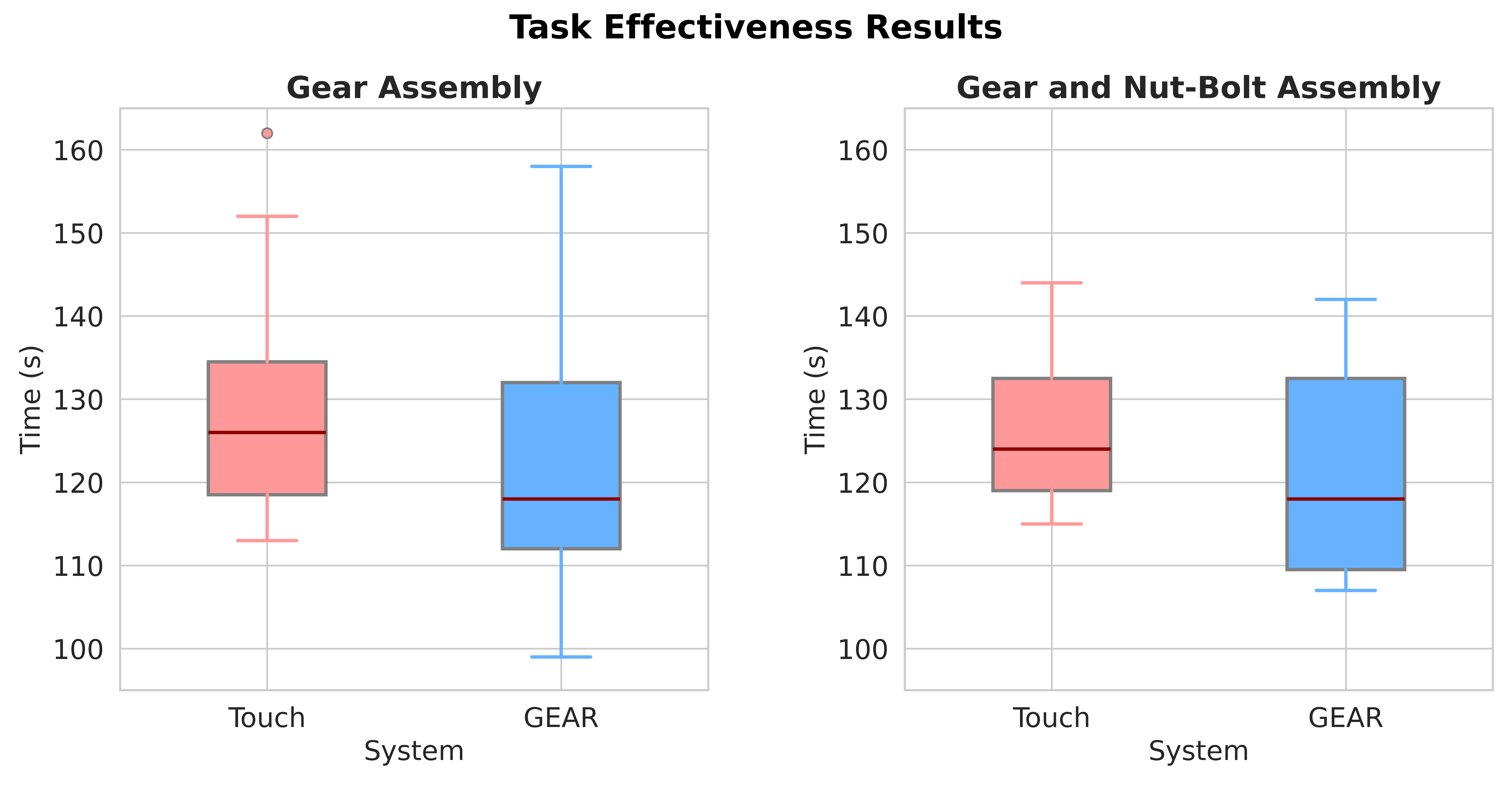}
    \caption{The figure presents two boxplots comparing task effectiveness, measured by time taken to accomplish tasks, for the two experiments: \textit{Gear Assembly} and \textit{Gear and Nut-Bolt Assembly}. In each plot, we analyze the distribution of completion times from two different groups of 15 users, each of whom interacted with both the touch interface and the GEAR.}
    \vspace{-3mm}
    \label{fig:time}
\end{figure}

\subsection{Workload Analysis (NASA-TLX)}
Participants' workload was assessed using the NASA-TLX Index, which evaluates cognitive, physical, and temporal demands, along with effort, performance, and frustration. Figure~\ref{fig:tlx} shows the results for both experiments, with mean scores in Table~\ref{tab:mean_tlx}. Although median scores were low for both interfaces, the touch interface exhibited greater variance in physical and temporal demands, performance, and effort, implying that some participants experienced higher workloads using the touch interface during the task. We also analyzed the results using a paired t-test to identify whether task complexity and interaction modality impact users' perceived workload and performance. The t-test results revealed no significant difference between touch and gaze for the \textit{Gear Assembly} task (t~=~1.6198, p~=~0.1276), but a significant difference for the \textit{Gear and Nut-Bolt Assembly} task (t~=~2.3294, p~=~0.0353), indicating that, at $\alpha~=~0.05$, interaction modality impacts workload more as task complexity increases. Across individual workload metrics, a few key differences were identified:

\begin{itemize}
    \item Physical demand showed a statistically significant difference, with touch interface being more physically demanding than GEAR in both tasks (p~$<$~0.05).
    \item Differences in performance and effort emerged as the task became more challenging. GEAR required less effort (p~$=$~0.0163) and yielded better performance (p~$=$~0.0006) in \textit{Gear and Nut-Bolt Assembly}. This is also evident in the boxplots in Figure~\ref{fig:tlx}, where the subscales for performance, effort, and physical demand were, on average, lower for the gaze interface.
    \item For other metrics (\textit{e.g.}, mental demand, temporal demand, frustration), no significant differences were found between touch interface and GEAR across both tasks.
\end{itemize}

This partially supports \textbf{H1}, indicating that GEAR reduces workload in terms of physical demand, performance, and effort, especially in complex tasks. 


\subsection{Usability Evaluation (SUS)}
To assess overall usability and user satisfaction, we employed the System Usability Scale (SUS). Figure~\ref{fig:sus} shows boxplots comparing the SUS scores for both interfaces. Both interfaces received high scores, indicating generally “good” usability with the standard touch interface receiving marginally higher ratings, although no statistical difference was observed (p~$>$~0.05). Participants reported that the touch interface felt more intuitive, leading to fewer interruptions and smoother interactions with the robot. Notably, in the more complex \textit{Gear and Nut-Bolt Assembly} task, the usability of the touch interface slightly decreased. While some users really appreciated the gaze interface, giving it the highest scores, others were more skeptical, likely due to the newness of the technology, their limited prior experience with eye trackers, and potential discomfort when not using corrective lenses during the task. Attaching vision correction lenses on the eye tracker for each individual user might alleviate discomfort. Overall, although GEAR demonstrated strong usability, the SUS results showed no statistical difference between the interfaces; therefore \textbf{H2} was not supported.

\subsection{Post-Experiment Interview}
In post-experiment interviews (see Section \ref{sec:questionnaire}), participants were asked about the perceived performance of both systems and their preferences. Notably, a total of 16 participants (8 in each experiment) preferred GEAR, while 14 (7 in each experiment) favored the touch interface. Those who preferred the touch interface highlighted its familiarity, stating: “Touch was more intuitive”, “I’m already used to it”, and “Glasses are uncomfortable”. Users who favored GEAR mentioned benefits such as: “Hands-free interaction”, “It was more efficient”, “I could focus more on the task”, and “I could directly select what I saw”.
Across both experiments, 19 users found object selection with GEAR easy or very easy, 6 found it normal, 4 found it difficult, and 1 very difficult. Regarding robot response time, 9 participants noted that the robot’s response after object selection felt slightly slow, while the rest found it appropriate. 

\subsection{Task Effectiveness}
To evaluate the task effectiveness, we analyzed the task completion time, since the error rate (\textit{i.e.}, the proportion of incorrectly selected components relative to the total requested) is negligible. The results are summarized in Figure~\ref{fig:time}.
On average, participants using GEAR completed tasks at a similar or faster pace compared to those using the touch-based interface. However, individual completion times varied, with some participants showing minimal differences, while others experienced longer durations due to usability challenges. 
A paired t-test for \textbf{H3} failed to reject the null hypothesis at conventional significance levels (p~$>$~0.05). Specifically, for the \textit{Gear Assembly} task, the p-value was 0.1322 (t~=~1.4862), and for the \textit{Gear and Nut-Bolt Assembly} task, it was 0.1594 (t~=~1.5987). Noteably, the lowest recorded task times among users were achieved with GEAR, while their performance with the touch interface was average (\textit{e.g.}, one participant finished the \textit{Gear Assembly} task in 98 seconds with GEAR compared to 135 seconds with touch, and another in 110 seconds with GEAR versus 122 seconds with touch). We interpret this result as an indication that the experiment’s complexity may not have been sufficient to highlight performance differences and that most users required more time to get familiar with using GEAR.
Crucially, the touch-based interface never significantly outperformed GEAR.
These results suggest that while GEAR has potential advantages, future studies should explore how user familiarity with the system and interaction flow influence task completion times.

\section{Discussion}
The results of our study indicate that GEAR improved human-robot collaborative assembly over the touch interface in key aspects such as workload and task effectiveness. However, in terms of perceived usability (SUS scores), both the gaze and touch systems were rated similarly, falling within the ``good" usability range. Participants found both interfaces easy to use, with touch likely benefiting from greater familiarity in everyday interactions rather than any inherent usability advantage. Even though the gaze-based system did not outperform the touch-based one, it was also not worse than it. This is despite the touch interface being much more familiar to participants from everyday interactions. This shows that using the gaze interface, although not familiar, was sufficiently natural and intuitive that it could compete with touch. On top of that, using the gaze-based interface partially lowered the workload. Arguably, the benefits would only increase once users get more familiar with using gaze, so we believe the proposed gaze interface could be an excellent replacement for the standard touch-based ones in HRC.

Some participants with corrected vision preferred the touch interface because they had to remove their glasses for the eye tracker to function properly, as we did not have special lenses for the eye tracker. This impacted their comfort and lowered usability ratings for GEAR. While the gaze-based interface improves performance, it requires more mental effort and introduces potential discomfort. The results suggest that incorporating gaze-based interaction into human-robot collaboration systems can significantly enhance efficiency, particularly in complex assembly tasks. However, further research is needed across different tasks of varying complexity to fully understand the potential and limitations of gaze-based systems. Future work should explore performance in more complex and realistic settings (\textit{e.g.}, denser object arrangements), where selection is more challenging. Further analysis of gaze-based selection errors might reveal links to eye-tracking accuracy or part size. In addition, diminished user control in gaze \cite{belardinelli2024gaze} may also affect the users' preferences.  

\section{Conclusion}
In this work, we developed GEAR, an intuitive system to investigate how gaze might enhance human-robot collaborative assembly. We employed 2D gaze matching from an eye tracker, paired with object detection on both the user's and robot's side, to accurately infer the user's intentions and translate them into robot commands. We performed a user study with 30 subjects to evaluate GEAR against a traditional touchscreen interface across two assembly tasks. From the experiments, we found several key insights. Eye gaze proved to be a valuable and effective signal for reducing workload, particularly in complex tasks during collaborative assembly. Despite its advantages, GEAR's perceived usability was similar to the touch interface, with nearly half of the participants favoring touch due to its familiarity from daily use. With increased familiarity, GEAR's benefits would likely become more pronounced, making the gaze interface a promising alternative to traditional touch-based systems in HRC.


\bibliographystyle{IEEEtran}
\bibliography{bibliography}
\end{document}